\documentclass[10pt,twocolumn,letterpaper]{article}

\usepackage{wacv}
\usepackage{times}
\usepackage{epsfig}
\usepackage{graphicx}
\usepackage{amsmath}
\usepackage{amssymb}
\usepackage{color,soul}
\DeclareMathAlphabet\mathbfcal{OMS}{cmsy}{b}{n}
\usepackage{url}
\usepackage{multirow}
\usepackage{makecell}
\usepackage{makecell}
\usepackage{enumitem}
\usepackage{arydshln}
\usepackage{nohyperref}

\wacvfinalcopy 

\def\wacvPaperID{1251} 

\ifwacvfinal\fi
\begin{document}

\title{Establishing a Strong Baseline for Cross-Domain Person ReID}

\author{Devinder Kumar*, \text{Parthipan Siva}$^\dag$, \text{Paul Marchwica}$^\dag$, \text{Alexander Wong*}\\
*University of Waterloo \hspace{2.4cm}  $^\dag$\text{SPORTLOGiQ Inc.,}\\
Waterloo, Ontario, Canada\hspace{1.6cm} Kitchener, Ontario, Canada\\
{\tt\small \{d22kumar,a28wong\}@uwaterloo.ca} \hspace{0.3cm} \tt\small \{parthipan,paul\}@sportlogiq.com}

\maketitle
\ifwacvfinal\thispagestyle{empty}\fi

\begin{abstract}
Person re-identification (ReID) is a challenging problem in computer vision, and critical for large-scale video surveillance scenarios where an individual could appear in different camera views at different times. Recently there has been an interest in cross-domain approaches for person ReID,
which leverages data from source domains that are different than the target domain. These approaches are more practical for large scale deployment without requiring on-site training or on-site manual annotation.
In this study, we take a systematic approach to establishing a large baseline source and target domain to act as a comprehensive benchmark for cross-domain person ReID. 
Furthermore, 
we establish a strong baseline method for cross-domain person ReID as a benchmark reference. Experiments show that a source domain composed of two of the largest person ReID domains (SYSU and MSMT) performs well across six commonly-used target domains. Furthermore, we show that two of the recent commonly-used target domains (PRID and GRID) have too few query images to provide meaningful insights. Based on our findings, we propose the following balanced baseline for cross-domain person ReID: i) a multi-target domain consisting of  Market-1501, DukeMTMC-reID, CUHK03, PRID, GRID and VIPeR and ii) a fixed split multi-source domain consisting of SYSU, MSMT, Airport and 3DPeS and iii) a leave-one-out multi-target domain consisting of SYSU, MSMT, Airport, 3DPeS, Market-1501, DukeMTMC-reID, and CUHK03 where Market-1501, DukeMTMC-reID, and CUHK03 are left out if testing on the same dataset.
\end{abstract}

\begin{figure}[t]
\centering
\includegraphics[width=0.49\textwidth]{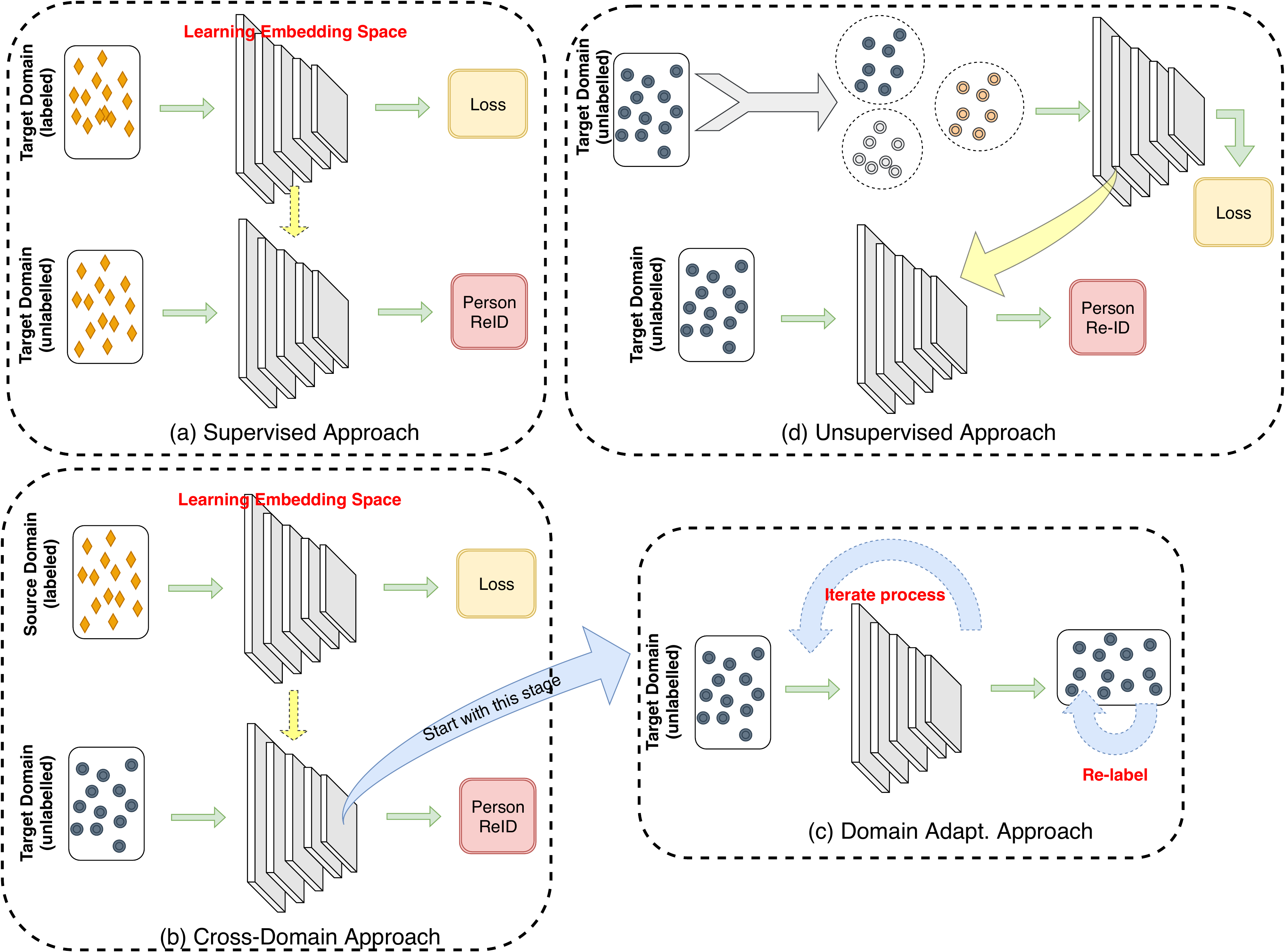}
\caption{Common learning approaches used in person ReID. (a) Supervised approaches train embedding space from labeled training data from the test environment (target domain). (b) Cross-domain approaches train embedding space from labeled training data from source domains that are independent of the target domain. (c) Domain Adaptation approaches adapt the cross-domain embedding space to target domain using unlabeled data from the target domain. (d) Unsupervised approaches learn embedding space directly from unlabeled target domain data without any external information.}
\label{fig:IDapproches}
\end{figure}

\section{Introduction}

One of the fundamental computer vision problems for the purpose of large-scale computerized video surveillance is the ability to identify an individual across a multitude of cameras in a given environment. This requires the ability to match an individual from one camera view to another, and is commonly referred to as the person re-identification (ReID) problem. One of the most promising approaches to person ReID that has achieved state-of-the-art results in recent years leverages deep convolutional neural networks (CNN) to learn an embedding space~\cite{zheng2015scalable,Zeng2018hierarchical,liu2017Stepwise,wang2018transferable}. Current research on person ReID using deep convolutional neural networks can be categorized as follows (Fig.~\ref{fig:IDapproches}):

\noindent \textbf{Supervised:} Supervised approaches explicitly assume that manually annotated data in the form of matched individuals across cameras is available for the camera network where person ReID is to be deployed (target domain). 


\noindent \textbf{Cross-Domain:} Cross-domain approaches assume the availability of manually annotated data from a single or multiple source domains which are different from the target domain. Embedding space is trained on the source domains, then applied to the target domain. 

\noindent \textbf{Domain Adaptation:} Domain adaptation approaches start with cross-domain result -- that is they start with an embedding space learned from a different source domain -- then utilize unlabeled data from the target domain to adapt to the target domain. 


\noindent \textbf{Unsupervised:} Unsupervised approaches do not use manually annotated data from the target domain or any source domains. They only utilize unlabeled data from the target domain to train the deep model's embedding space. 


Supervised approaches are the most extensively studied approaches in person ReID~\cite{zhang2019densely,tay2019aanet,zheng2019pyramidal,zheng2019joint} and achieve the best results so far. However, from a practical point of view this approach is infeasible for large scale deployment of person ReID because of the need for manually annotated data from every target domain. On the opposite end of the manual annotation spectrum are unsupervised approaches where no manual annotation is needed at all. This is a promising avenue of research with some amazing results already reported~\cite{song2018unsupervised,li2018unsupervised,qin2015unsupervised,fan2018unsupervised}. However, this approach assumes the availability of training hardware at the target sites or the ability to transfer unlabeled data from target site to a training facility. Again, this complicates large scale deployment of person ReID using such an approach. Furthermore, the system requires a learning period in which person ReID functionality will not be available at all. Domain adaptation is a compromise between supervised and unsupervised approaches but it still requires on-site training.

Cross-domain approaches can be regarded as the ideal approaches for practical deployment because a pre-trained model from annotated source domain(s) can simply be deployed to any target domain without on-site training. Currently, cross-domain research in person ReID is split into two key directions, with some works reporting cross-domain results as part of domain adaptation research \cite{li2018adaptation}, while other works looking directly at the cross-domain problem \cite{jia2019frustratingly,song2019generalizable}. 

One of the biggest challenges impeding advancements in the deployment and development of cross-domain approaches is the lack of consistency in the use of source domains and target domains for training and evaluation purposes.  This challenge makes it very difficult to compare existing methods to determine the strengths and limitations of methods and identify strong approaches for deployment and further development, and difficult to judge the improvements one would get when extending upon fundamental ideas on cross-domain approaches.  Many papers \cite{li2014deepreid,lisanti2015person,li2018unsupervised} consider single source domain and single target domain scenarios. This results in easier comparison between different approaches but does not take advantage of the existing multiple person ReID datasets. Recently there has been an interest in cross-domain person ReID using large diverse source domain~\cite{jia2019frustratingly,marchwica2018evaluation,song2019generalizable}. Existing works are split on how to create a large diverse source domain. Some papers use synthetic data generation~\cite{bak2018domain} while the majority of other papers seek to combine existing person ReID datasets into a larger diverse source domain~\cite{jia2019frustratingly,marchwica2018evaluation}. Combining existing datasets is a suitable approach however all existing papers use different combinations of datasets, making it very difficult to benchmark and understand the relative performance between different methods. In addition, most works look at only a few datasets in the source domain~\cite{jia2019frustratingly,marchwica2018evaluation}. One exception is~\cite{song2019generalizable}, however their target domain is limited to some very small datasets. In this paper we aim to establish a strong baseline for benchmarking cross-domain person ReID by looking at:
\begin{enumerate}
    \item a fixed source and target domain split that has large varied datasets in both source and target domains
    \item a leave-one-out source and target domain split that maximizes the source domain size. 
\end{enumerate} 


We take a systematic approach to establishing a large baseline source and target domain to act as a comprehensive benchmark for cross-domain person ReID. We accomplish this by conducting a comprehensive analysis to study the similarities between source domains proposed in literature, and studying the effects of incremental increasing the size of the source domain. This allows us to establish a balanced source and target domain split that promotes variety in both source and target domains. Furthermore, using lessons learned from state-of-the-art supervised person ReID methods, we establish a strong baseline method as a benchmark reference for cross-domain person ReID.  We show, with such a large diverse source domain, a very strong performance can be achieved for cross-domain person ReID.

In summary, the key contributions of this study are as follows:
\begin{itemize}
\item conduct a comprehensive analysis to study the similarities between source domains proposed in literature,
\item study the effects of incrementally increasing the size of the source domain,
\item establish a large, baseline source and target domain based on the findings of the above analysis to act as a comprehensive benchmark for cross-domain person ReID.
\item establish a strong baseline method for cross-domain person ReID as a benchmark reference using lessons learned from the state-of-the-art supervised person ReID methods. 
\end{itemize}

\begin{figure*}[h!]
\centering
\includegraphics[width=0.85\textwidth]{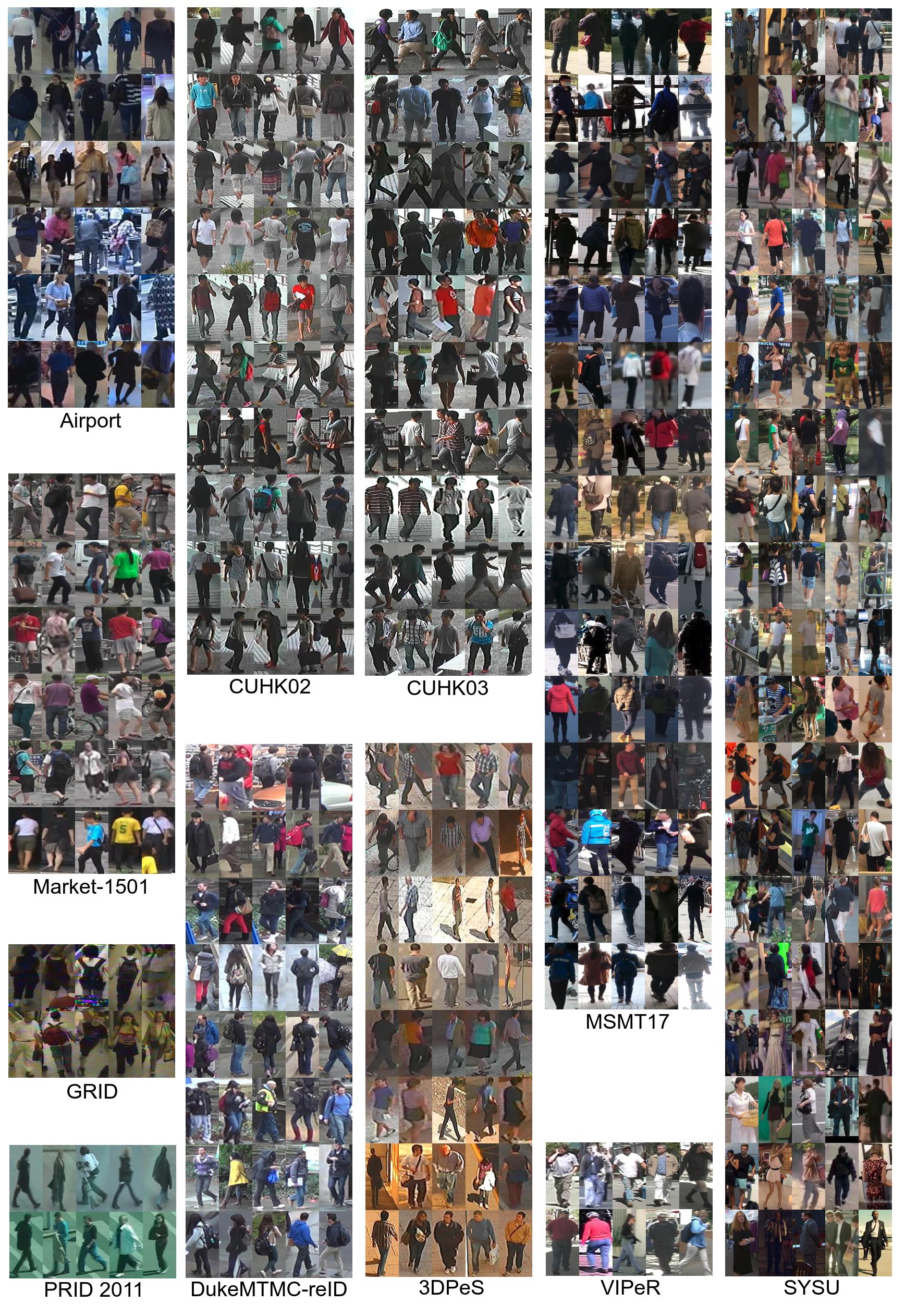}
\caption{Existing ReID datasets [Airport~\cite{karanam2018airport},  
				DukeMTMC-reID~\cite{zheng2017unlabeled},  
				Market-1501~\cite{zheng2015scalable}, 
			    CUHK03~\cite{li2014deepreid}, 
				SYSU~\cite{DBLP:journals/corr/XiaoLWLW16},
				MSMT17~\cite{Wei2018GAN},  
				CUHK02~\cite{li2013locally}, 
				PRID2011~\cite{PRID2011Dataset},  
				GRID~\cite{GRIDDataset},  
				VIPeR~\cite{ViperDataset}, 
				3DPeS~\cite{baltieri2011_308}].
 All images have been re-sized to 256x128 for easier comparison.}
\label{fig:datasets}
\end{figure*}

\begin{table}[t]
	\caption{List of different source and target domain datasets.}
	\label{TAB:DatasetComposition}
	\begin{center}
		\setlength{\tabcolsep}{0.2cm}
		\scriptsize{
			\begin{tabular}{l||c|c|c|c}
				\hline
				Dataset  & \# IDs & \# Images & \# Cams & Common Test-Set\\
				\hline\hline

				Airport~\cite{karanam2018airport}  & 1381 & 8660 & 6 & -- \\
				[1pt]
				DukeMTMC-reID~\cite{zheng2017unlabeled}  & 1404 & 32948 & 8 & \checkmark \\
				[1pt]
				Market-1501~\cite{zheng2015scalable} & 1501 & 32668 & 6 & \checkmark \\
			    [1pt]
			    CUHK03~\cite{li2014deepreid}  & 1467 & 14097 & 10 & \checkmark\\
				[1pt]
				SYSU~\cite{DBLP:journals/corr/XiaoLWLW16} & 11934 & 34574 & N/A & --\\
				[1pt]
				MSMT17~\cite{Wei2018GAN}  & 3060 & 126142 & 15 & --\\
				[1pt]
				CUHK02~\cite{li2013locally}  & 1816 & 7264 & 10 & -- \\
				[1pt]
				PRID2011~\cite{PRID2011Dataset}  & 385 & 1134 & 2 & \checkmark\\
				[1pt]
				GRID~\cite{GRIDDataset}  & 250 & 500 & 2 & \checkmark\\
				[1pt]
				VIPeR~\cite{ViperDataset}  & 632& 1264 & 2 & \checkmark \\
				[1pt]
				3DPeS~\cite{baltieri2011_308}  & 164 & 951 & 8 & --\\ 
				\hline
			\end{tabular}
			Note the counts presented here are for IDs that appear in multiple cameras. During training, only IDs that appear in multiple cameras were used. During test, standard splits and IDs were used. 
		}
	\end{center}
\vspace*{-2mm}
\end{table}


\section{Background}

The earliest reported results for cross-domain person ReID are from domain adaptation papers \cite{li2014deepreid} that report ``direct transfer'' results before presenting the improvement due to the domain adaptation method. These papers typically used a single source and target domain, for example Market-1501 as source domain and DukeMTMC-reID as target domain \cite{li2014deepreid,lisanti2015person}. Following in the footsteps of these papers, most methods that looked at cross-domain person ReID also used a single source and target domain \cite{liu2019attention}. As a result these approaches do not take advantage of variation in source domain that can be achieved by combining data collected from different source domains. 

Some early works on direct transfer, for example CAMEL \cite{yu2017cross}, combined a few datasets into a multi-source domain and tested on a few target domain datasets. Since CAMEL \cite{yu2017cross}, very few works has combined datasets to form a diverse source domain until~\cite{marchwica2018evaluation,jia2019frustratingly,song2019generalizable}. The recent work~\cite{jia2019frustratingly,song2019generalizable} formed a large diverse source domain but their focus was on a new training mechanism and as such baseline results were not fully investigated. Furthermore, their target domains (test sets) were limited to only few small datasets. Hence, the need to study the performance of diverse source domain on large target domains.

\section{Methodology}

The goal of this study is to:
\begin{enumerate}
    \item Investigate and evaluate the similarity between previously proposed source domains.
    \item Explore the effects of incrementally increasing the source domain.
    \item Systematically establish a large source and target domain for cross-domain person ReID by combining multiple existing datasets.
    \item Establish a strong baseline method for cross-domain person ReID based on lessons learned from past methods in literature.
\end{enumerate}

\noindent To this end, we look at the existing datasets for person ReID and establish a set of source domains and target domains that are commonly used across existing literature. We perform person ReID using a strong baseline method learned from previous state-of-the-art techniques in supervised person ReID to enable consistent evaluation and analysis, and study the effects of the source domain variation on cross-domain person ReID.


\begin{table*}[t]
	\centering
	\setlength{\tabcolsep}{0.2cm}
	\caption{Cross-domain performance using single source domain. 
		$1^\text{st}$/$2^\text{nd}$ best results are in \textbf{\color{red}red}/\textbf{\color{blue}blue}.}
	\label{tab:crossdomainsingledataset}
	\small{
	\begin{tabular}
		{p{2.2cm}||c c||c c||c c||c c||c c||c c||c c}
		\hline
		\multirow{2}{*}{Source $\downarrow$ / Target $\rightarrow$}
		& \multicolumn{2}{c||}{\makecell{Market- \\1501}} 
		& \multicolumn{2}{c||}{\makecell{Duke \\ MTMC-reID}}
		& \multicolumn{2}{c||}{CUHK03}       
		& \multicolumn{2}{c||}{PRID\cite{PRID2011Dataset}} 
		& \multicolumn{2}{c||}{VIPeR\cite{ViperDataset}} 
		& \multicolumn{2}{c||}{GRID\cite{GRIDDataset}}
		& \multicolumn{2}{c}{Avg.}
		\\ \cline{2-15}
		& R1	& mAP  & R1	& mAP	& R1	& mAP 	& R1  & mAP	 	& R1   & mAP   & R1  & mAP & R1 & mAP
		\\ \hline \hline
		
		\makecell{Market- \\ 1501}\cite{zheng2015scalable}  & --         & -- & 43.6	 & 24.3    	& 6.3	 & 5.2   & \color{blue}23.0 & \color{blue}27.1 & 32.8 & 37.0 & 19.4 & 23.1 & -- & --  
		\\ \hline
		\makecell{Duke \\ MTMC-reID}\cite{zheng2017unlabeled} &56.2       & 26.5 	& --	 & --     &  5.2	 & 4.8  & 20.4 & 25.0 & 31.2 & 35.3 & 11.5 & 14.6 & -- & --
		\\ \hline
		CUHK03\cite{li2014deepreid}    & 56.9       & 30.8    & 33.1	 & 16.9    	& --	 & --   & 14.4  & 17.6 & 29.9 & 34.8 & 21.8  & 25.3 & -- & --
		\\ \hline
		Airport\cite{karanam2018airport}   & 51.1       & 25.3    & 33.8	 & 17.4    	& 6.0	 & 5.4  & 17.2 & 20.6  & 25.3 & 30.5   & \color{blue}28.2 &  \color{blue}32.1 & 26.9 & 21.9
		\\ \hline
		SYSU\cite{DBLP:journals/corr/XiaoLWLW16}  & \color{red}65.2       & \color{red}40.0    & \color{blue}42.9	 & \color{blue}27.2    	& \color{blue}8.8	 & \color{blue} 7.4   & \color{red}41.5 & \color{red}47.1 & \color{blue} 37.7  & \color{blue}42.0 & \color{red}28.5 & \color{red}32.8  & 37.4 & 32.8
		\\ \hline
		MSMT\cite{wei2018person}   & \color{blue}64.8       & \color{blue}36.6      & \color{red}64.5	 & \color{red}43.3   	& \color{red}16.4	 & \color{red}14.2   & 21.9 & 25.8& \color{red}44.1 & \color{red}48.7  & 6.5 & 8.9 & 36.4 & 29.6
		\\ \hline
		CUHK02\cite{li2013locally}    & 53.5       & 24.9    & 39.9	 & 21.5    	& 36.2*	 & 31*   & 16.4 & 20.4 & 29.5 & 34.1   & 16.2 & 19.3 & 32.0 & 25.2
		\\
		\hline 
	\end{tabular}
	\text{* Not considered as fair cross-domain test as environment is the same}
	}
\end{table*}


\subsection{Domain analysis}

In order to establish a baseline source and target domain for cross-domain person ReID in a systematic manner, we need to first study the effects of the differences in source and target domains as well as the effect of diversity in source domain. In Fig.~\ref{fig:datasets} and Table~\ref{TAB:DatasetComposition}, we list the existing domains for person ReID with more than 100 unique persons in the dataset.\footnote{While all efforts were made to include all domains with sufficient unique IDs, several lesser known domains (SAIVT-SoftBio, PKU-Reid and RPIfield) were omitted} We do not use any domains with data overlaps such as CUHK02 and CUHK01, or MARS and Market-1501, etc. One exception is that we include CUHK02 and CUHK03 in our analysis as they have no overlaps in people who appear in the domains but they were obtained in the same environment. As this is a study regarding cross-domain analysis, we cannot use CUHK02 in training and CUHK03 in testing (or vice versa) without negatively impacting the study.

For each of the selected domains we limit our training to person IDs that are present in more than one camera views because our goal is person ReID across camera views. When including a dataset in the source domain all IDs from the dataset (including test IDs) are used for training. 

Of the datasets in Table~\ref{TAB:DatasetComposition}, DukeMTMC-reID, Market-1501 and CUHK03 are commonly used large target domains for domain adaptation \cite{zheng2017unlabeled} and as such also have various reported results for cross-domain person ReID. Furthermore, PRID 2011, GRID and VIPeR are small domains which have been recently used for testing cross-domain person ReID \cite{jia2019frustratingly,song2019generalizable}. As such, we use these six datasets in the target domain for testing cross-domain person ReID. Of these domains, DukeMTMC-reID~\cite{zheng2017unlabeled}, Market-1501~\cite{zheng2015scalable} and CUHK03~\cite{li2014deepreid} have pre-established test probe and test gallery splits which we use for our test. For the smaller domains (PRID 2011, GRID and VIPeR), we use the standard 10 random splits as in \cite{jia2019frustratingly} for obtaining Rank-1 and mAP results.


\subsection{Baseline method analysis} 
Next, we aim to establish a strong baseline method for cross-domain person ReID to facilitate the evaluation over different domains based on lessons from state-of-the-art literature. As the starting point, ResNet-50, pre-trained on ImageNet, was used as the baseline architecture as this particular architecture has been shown to be the most widely used and effective architecture for obtaining a strong baseline and improved results when used alongside either ID/softmax loss or triplet-based embedding spaces in recent person ReID methods~\cite{li2018unsupervised,zheng2015scalable}. Recently, Luo et al.~\cite{luo2019bag} have established a strong baseline for supervised person ReID by using a combination of ID/softmax loss and triplet loss along with certain network architecture changes (such as Batch Normalization Neck (BNNeck)) and \textit{tricks}. Using these modifications, they were able to produce state-of-the art baseline performance for supervised person ReID for various single source domain based ReID. In~\cite{luo2019bag}, the authors also experimented on cross-domain ReID using single source and single target domains, but did not present extensive cross-domain results. They did find that some of the tricks such as random erasing don't work as effectively for cross-domain as they do for supervised ReID. Similarly, normalizing features using batch normalization at the end of the network have been recently found to be successful for obtaining impressive performance for person ReID~\cite{figueira2014hdaDataset, jia2019frustratingly}. Hence, we incorporated the aforementioned lessons learned in supervised approaches, as well as incorporating additional best practices such as BNNeck, no random erasing and batch normalization at the end into our ResNet50 baseline architecture. 

\section{Experiments and discussions}

Using the datasets established in Table~\ref{TAB:DatasetComposition} we study the effects of source and target domain similarity, the effects of incrementally increasing the source domain size and finally compare the multi-source domain baseline results presented here to existing cross-domain results.

\subsection{Source-target domain similarity}

To investigate which domains are similar to each other, we perform cross-domain person ReID by using each dataset with more than two camera views as the source domain. The target domains are the aforementioned six datasets (Table~\ref{TAB:DatasetComposition}). 
The result of this can be found in Table~\ref{tab:crossdomainsingledataset}. 

Based on Table~\ref{tab:crossdomainsingledataset}, it can be observed that SYSU and MSMT as source domains perform fairly well across most target domains. This is not surprising considering that these two domains have the largest number of unique individuals appearing in multiple cameras and also the largest number of cameras in their camera networks \footnote{SYSU used data from television shows and photos taken with handheld digital cameras as people were walking around the city}. 

Most target domains performed well across all source domains with the exception of CUHK03. It can be seen that ReID performance for CUHK03 is low compared to other target domains except when CUHK02 is used as the source domain. However, as stated before the environment used for CUHK02 and CUHK03 are very similar (Fig.~\ref{fig:datasets}) and as such this can't be considered a fair cross-domain person ReID setting.

\begin{figure}[ht]
\centering
\includegraphics[width=0.48\textwidth]{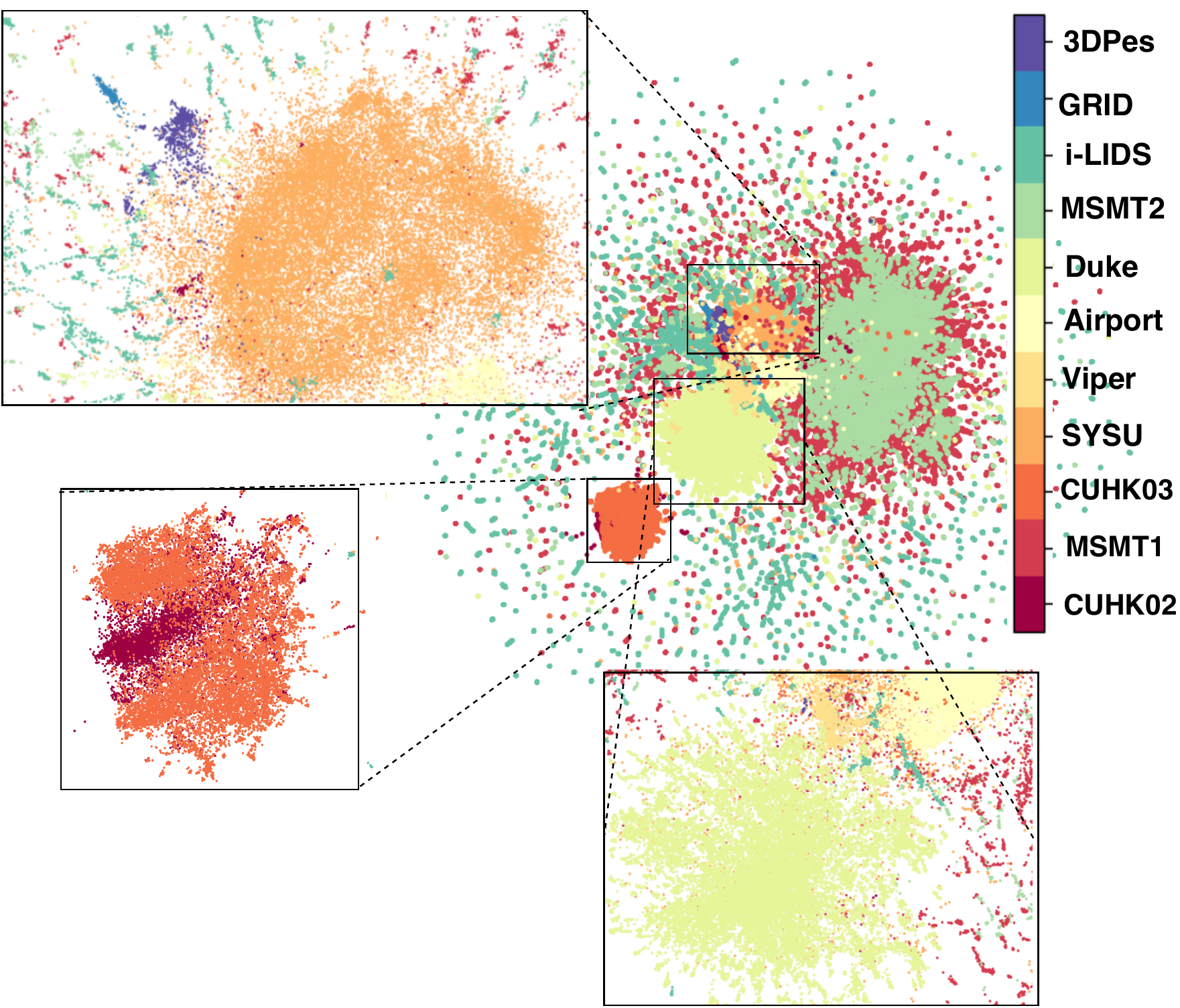}
\caption{UMAP for various person ReID domains from embeddings obtained for pre-trained ResNet50 model. Closeness in the embedding space shows the similarity between the various person ReID domains. MSMT1 and 2 are MSMT17 train and val-sets.}
\label{fig:umap}
\end{figure}

\noindent \textbf{UMAP Illustration}: To further illustrate the similarity of various person ReID domains, we created a Uniform Manifold Approximation and Projection (UMAP)~\cite{mcinnes2018umap} illustration as shown in Fig.~\ref{fig:umap}. We created the UMAP figure using embeddings from various domains obtained from the proposed baseline method for cross-domain ReID. From the figure, it can be observed that CUHK02 and CUHK03 are very similar and furthest from the rest of the domains including MSMT and SYSU, thus explaining the poor performance of various domains on CUHK03. Also, SYSU is closest to GRID, hence it performs better than MSMT on GRID.



\begin{table*}[!htbp]
	\centering
	\setlength{\tabcolsep}{0.2cm}
	\caption{Cross-domain performance using increasingly larger source domain compared to other approaches to ReID. \newline S: SYSU, MT: MSMT17, C03: CUHK03, C02: CUHK02, C01: CUHK01, A: Airport, D: DukeMTMC-reID, M: Market-1501, ex: MARS-M, 3D: 3DPeS, G: GRID, V: VIPeR, iL: iLDS. Best results for each section is in \textbf{\color{red}red}. }
	\label{tab:incrementalcrossdomain}
	\small{
	\begin{tabular}
		{p{3.5cm}||c c||c c||c c||c c||c c||c c}
		\hline
		\multirow{2}{*}{Source $\downarrow$ / Target $\rightarrow$}
		& \multicolumn{2}{c||}{\makecell{Market- \\1501}\cite{zheng2015scalable}} 
		& \multicolumn{2}{c||}{\makecell{DukeMTMC \\-reID}\cite{zheng2017unlabeled}}
		& \multicolumn{2}{c||}{CUHK03\cite{li2014deepreid}} 
		& \multicolumn{2}{c||}{PRID\cite{PRID2011Dataset}}
		& \multicolumn{2}{c||}{VIPeR\cite{ViperDataset}} 
		& \multicolumn{2}{c}{GRID\cite{GRIDDataset}} 
		\\ \cline{2-13}
			& R1	& mAP  & R1	& mAP   & R1	& mAP   & R1  & mAP   & R1	& mAP   & R1 & mAP 
		\\ \hline \hline
		\textbf{Supr. Methods-SOTA} &  &    &  &  &  &  &  & &   &  &  &  \\
		 DSA~\cite{zhang2019densely} & \color{red}95.7 & 87.6   & -- & -- & 78.2 & 73.1 & -- & --& --  & -- & --& -- \\
		 Pyramid~\cite{zheng2019pyramidal} & \color{red}95.7 & \color{red}88.2   & -- & -- & \color{red}78.9 & \color{red}74.8 & -- & --& --  & -- & --& -- \\
		 AANet~\cite{tay2019aanet}  & -- & --   & \color{red}90.4 & \color{red}86.8 & -- & -- & -- & --& --  & -- & --& -- \\
		\hline 		\hline
		\textbf{Unsupervised Methods} &  &    &  &  &  &  &  & &   &  &  &  \\
		 ARN~\cite{li2018adaptation} & \color{red}70.3 & 39.4   & 60.2 & 33.4 & -- & -- & -- & --& --  & -- & --& -- \\
		 EANet~\cite{huang2018eanet} & 66.4 & 40.6   & 45.0 & 26.4 & 51.4 & 31.7 & -- & --& --  & -- & --& -- \\
		 ECN~\cite{zhong2019invariance} & 75.1 & 43.0   & 63.3 & 40.4 & -- & -- & -- & --& --  & -- & --& -- \\
		 DPRID~\cite{wu2019distilled} & -- & --   & 48.4 & 29.4 & -- & -- & -- & --& --  & -- & --& -- \\
		 MAR~\cite{yu2019unsupervised} &  67.7 & 40.0   & \color{red}87.1 & \color{red}48.0 & -- & -- & -- & --& --  & -- & --& -- \\
		 TAUDL~\cite{li2018unsupervised} & 63.7 & 41.2   & 61.7 & 43.5 & 44.7 & 31.2 & 49.4  & --& --  & -- & --& -- \\
		 UTAL~\cite{li2019unsupervised} &69.2 & \color{red}46.2 & 62.3 & 44.6 &\color{red}56.3 &\color{red}42.3 & \color{red}54.7& --  & -- & --& --& --\\
		\hline 		\hline
		\textbf{Domain Adapt. Methods} &  &    &  &  &  &  &  & &   &  &  &  \\ 
		HHL(D,M)~\cite{zhong2018generalizing} & 62.2 & 31.4 & 46.9  & 27.2  & -- & -- & -- & --& --  & -- & --& -- \\
		HHL (C03)~\cite{zhong2018generalizing} & 56.8 &29.8 & 42.7  & 23.4  & -- & -- & -- & --& --  & -- & --& -- \\
		DAAN (C03,M)~\cite{xu2019cross} & 84.5 & -- & --  & --  & \color{red}78.2 & -- & -- & --& --  & -- & --& -- \\
		ATNet (D,M)~\cite{liu2019adaptive} & 55.7 & 25.6   & 45.1 & 24.9 & -- & -- & -- & --& --  & -- & --& -- \\
		CSGLP (D,M)~\cite{ren2019domain} & 63.7 & 33.9   & 56.1 & 36.0 & -- & -- & -- & --& --  & -- & --& -- \\
		ISSDA (D,M)~\cite{tang2019unsupervised} & \color{red}81.3 &\color{red} 63.1   & \color{red}72.8 & \color{red}54.1 & -- & -- & -- & --& --  & -- & --& -- \\
		\hline 		\hline
	    \textbf{Cross-Domain Methods} &  &    &  &  &  &  &  & &   &  &  &  \\
	    \textit{Single Source} &  &    &  &  &  &  &  & &   &  &  & \\
	    EANet (C03)~\cite{huang2018eanet} & 59.4 & 33.3   & 39.3 & 22.0 & -- & -- & -- & --& --  & -- & --& -- \\
	    EANet (D,M)~\cite{huang2018eanet} & 61.7 & 32.9   & 51.4 & 31.7 & -- & -- & -- & --& --  & -- & --& -- \\
	    DPAN (C03,M)~\cite{xu2019cross} & 56.0 & --   & -- & -- & \color{red}50.4 & -- & -- & --& --  & -- & --& -- \\
	    SPGAN (D,M)~\cite{deng2018image}  & 43.1 & 17.0   & 33.1 & 16.7 & -- & -- & -- & --& --  & -- & --& -- \\
	    DAAM (D,M)~\cite{Huang2019domain} & 42.3 & 17.5   & 29.3 & 14.5 & -- & -- & -- & --& --  & -- & --& -- \\
	    AF3 (D,M)~\cite{liu2019attention} & 67.2 & 36.3   & 56.8 & 37.4 & -- & -- & -- & --& --  & -- & --& -- \\
	    AF3 (MT)~\cite{liu2019attention} & 68.0 & 37.7   & 66.3 & 46.2 & -- & -- & -- & --& --  & -- & --& -- \\
	    PAUL (MT)~\cite{yang2019patch} & 68.5 & 40.1   & \color{red}72.0 & \color{red}53.2 & -- & -- & -- & --& --  & -- & --& -- \\ \hdashline
	    \textbf{\textit{Multi-Source}} &  &    &  &  &  &  &  & &   &  &  & \\
	    EMTL~\cite{xian2018enhanced} (C02+D+M) & 52.8 & 25.1   & 39.7 & 22.3 & -- & -- & -- & --& --  & -- & --& -- \\
	    GDIM(C02+03+D+M)~\cite{song2019generalizable} & -- & --   & -- & -- & -- & -- & 39.2 & 51.9 & 51.2  & 60.1 & 29.3 & 41.1   \\
	    DualNorm{\tiny (M,D,C02-03,S)} \cite{jia2019frustratingly} & -- & --   & -- & -- & -- & -- & \color{red}60.4 & --  & 53.9& -- & \color{red}41.4  & --  \\
	    CAMEL\tiny(V+C01+C03+S+M+exM)~\cite{yu2017cross} & -- & --   & -- & -- & \color{red}31.9 & -- & -- & --& -- & 30.9 & -- & --   \\
	    CAMEL \tiny (V+C01+C03+3D+iL)~\cite{yu2017cross} & 54.5 & 26.3   & -- & -- & -- & -- & -- & --& --  & -- & --& -- \\
		\hline
		\textbf{Ours:Cross-Domain} &  &    &  &  &  &  &  & &   &  &  &  \\ 
		S           & 65.2 & 40.0   & 42.9 & 27.2 & 8.8 & 7.4 & 41.5 & 47.1 & 37.7 & 42.0 & 28.5 & 32.8  \\
		S+MT        & 73.5 & 47.3 & 61.3 & 39.5 & 19.7 & 17.8 & 43.6 & 48.4 & 37.8 & 42.4 & 32.2 & 37.4   \\
		S+MT+A    & 76.7 & 51.5   & 63.2 & 42.0  & 22.8 & 21.1 & 50.6 & \color{red}55.7 & 52.8  & 57.2 & 35.0 & 39.8  \\		
		$\textbf{\color{cyan}S+MT+A+3D}$    & 75.9 & 51.9   & 63.6 & 42.7 & 23.6 & 21.4 & 40.9 & 46.3 & 55.7 & 59.8 & 38.0 & 42.6 \\
		$\textbf{\color{blue}S+MT+A+3D+C03+C02+D }$   & \color{red}80.5 & \color{red}56.8   & -- & -- & -- & -- & 46.5 & 51.6 & 57.2 & 61.2 & 39.4 & \color{red}43.0 \\
		$\textbf{\color{blue}S+MT+A+3D+C03+C02+M }$   & -- & --   & \color{red}67.4 & \color{red}46.9  & -- & -- & 46.9 & 52.5 & 59.0  & 62.8 & 33.4 & 39.0  \\
		$\textbf{\color{blue}S+MT+A+3D+D+M }$   & -- & --   & -- & --  & 29.4 & \color{red}27.4 & 43.7 & 49.4 & 59.4  & 63.2 & 34.2 &  38.8 \\
		$\textbf{\color{blue}S+MT+A+3D+\scriptsize{C03+C02+D+M}}$   & -- & --   & -- & -- & -- & -- & 43.7 & 49.4 &  \color{red}59.5 &  \color{red}63.5 & 34.4 & 39.0 \\
		\hdashline
		Max single source domain* & 65.2 & 40.0 & 64.5 & 43.3 & 16.4 & 14.2 & 41.5 & 47.1 & 44.1 & 48.7 & 28.5 & 32.8 \\
		Largest multi-source domain** & 80.5 & 56.8 & 67.4 & 46.9 & 29.4 & 27.4 & 43.7 & 49.4 &  59.5 &  63.5 & 34.4 & 39.0 \\
		\hline
	\end{tabular}
	}
	* Max along Table~\ref{tab:crossdomainsingledataset} columns.
	
	** Based on leave-one-out split, \textbf{\color{blue}blue} rows.
	
	Recommended fixed and leave-one-out split for cross-domain preson ReID in \textbf{\color{cyan}cyan} and \textbf{\color{blue}blue} respectively. 
	Text in \textbf{\color{red}red} indicates best R1 and mAP result for each method and target dataset.
\end{table*}


\begin{figure}[ht]
\centering
\includegraphics[width=0.48\textwidth]{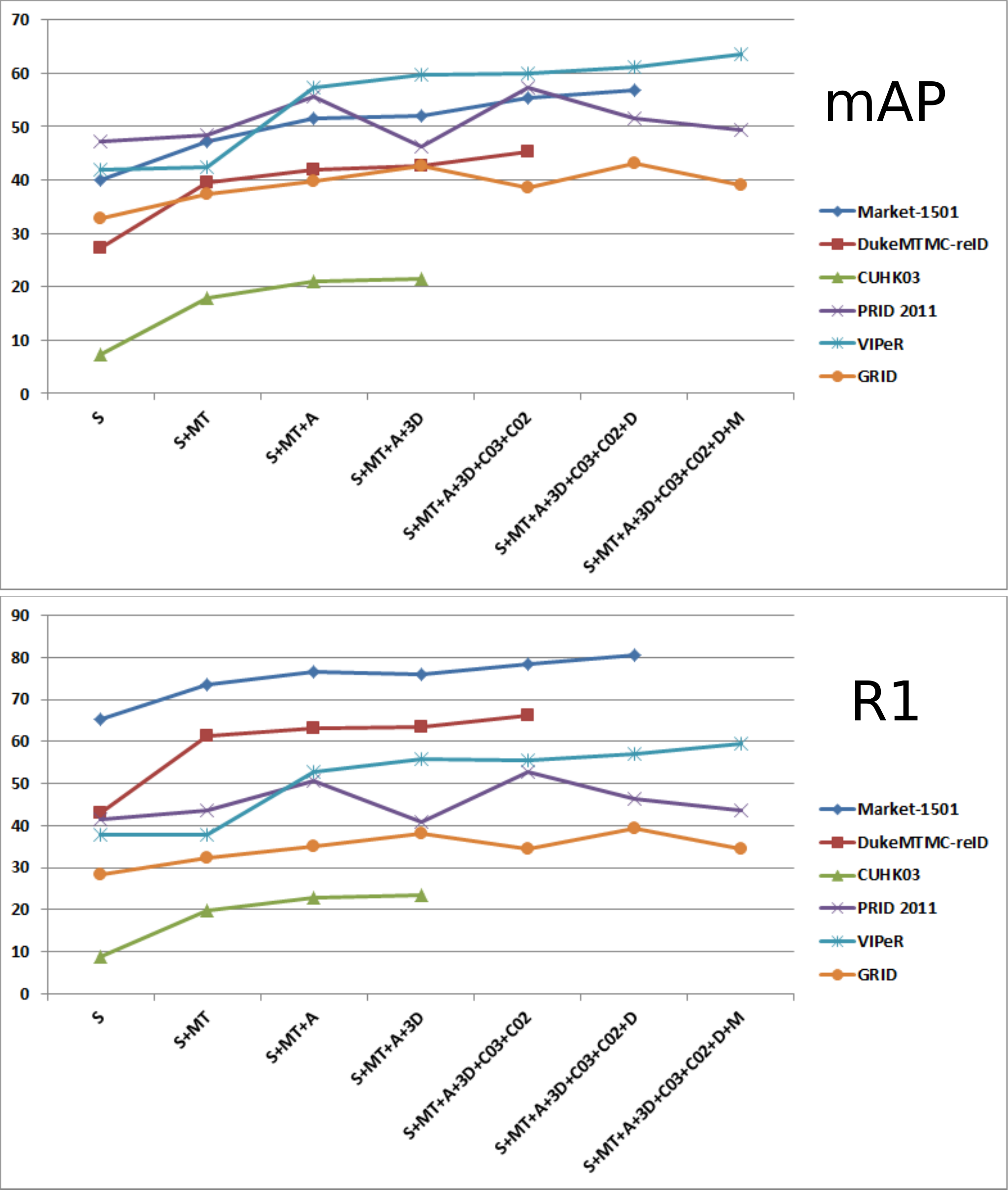}
\caption{Plot showing the performance of different person ReID domains on various combination of multi-source domain using mAP and Rank-1 (R1) as evaluation metric. S: SYSU, MT: MSMT17, A: Airport, 3D: 3DPeS, C03: CUHK03, C02: CUHK02, D: DukeMTMC-reID.}
\label{fig:plot}
\end{figure}

\subsection{Effects of increasing source domain size}

Starting with SYSU, the best average Rank-1 and mAP source domain across all target domains as seen in Table~\ref{tab:crossdomainsingledataset}, we progressively add source domains to see the effect on person ReID results on all target domains. The results are shown in Fig.~\ref{fig:plot} and Table~\ref{tab:incrementalcrossdomain}.

Looking at Fig.~\ref{fig:plot}, all target domains with the exception of GRID and PRID2011 show a gradual increase in performance (both Rank-1 and mAP) as we add more datasets to the source domain. This is what we would expect to see as we increase the source domain complexity (a.k.a. training set). The noisy performance of GRID and PRID2011 as source domain increases can be associated to the number identities in these two dataset. As seen from Table~\ref{TAB:DatasetComposition}, GRID and PRID2011 have the least number of identities of the six target domains, making it hard to draw conclusions from these datasets' performance. This is a justification for the need to include larger target domain datasets for evaluation in order to obtain more generalisable conclusions from the test results.

\subsection{Is a larger source domain always best?}

If we look at the result of using the largest possible source domain per target domain vs. the maximum performance possible with single source domain (Table~\ref{tab:incrementalcrossdomain}), we see there is clearly a performance gain with the largest possible source domain. Of particular importance is the CUHK03 target domain results, where the multi-source domain nearly doubles the maximum performance by the best single source domain. This suggests simply having a very robust source domain composed from multiple environment conditions and a large number of unique individuals can obtain a very robust cross-domain person ReID results.

One important caveat can be learned by considering the DukeMTMC-reID performance. For DukeMTMC-reID, using MSMT dataset as the source domain we obtain a Rank-1 result of 64.5 and mAP of 43.3. This is fairly high performance relative to all other source domains considered in Table~\ref{tab:crossdomainsingledataset}, meaning there is a high similarity between MSMT and DukeMTMC-reID datasets. When combining MSMT with SYSU as source domain we see a drop in performance: Rank-1 result of 61.3 (-3.2) and mAP of 39.5 (-3.8). In fact even when using source domain composed of MSMT, SYSU, Airport, and 3DPeS there is a drop in performance: Rank-1 result of 63.6 (-0.9) and mAP of 42.7 (-0.6). Only with the addition of CUHK03 and CUHK02 datasets does the performance increases.

As a result, of the DukeMTMC-reID performance, it is worth noting that training with a few source domains that have high similarity to the target domain could result in better performance than having a large varied source domain. However, if deploying to multiple target domains then a large varied source domain is the best option.

\subsection{Recommended source domain splits}

Based on our study we recommend the use of a source domain composed of SYSU, MSMT, Airport and 3DPeS and target domain composed of Market-1501, DukeMTMC-reID, CUHK03, PRID, GRID and VIPeR for a balanced fixed source and target domain splits. The baseline performance of this split is shown in \textbf{\color{cyan}cyan} in Table~\ref{tab:incrementalcrossdomain}.

In order to test the performance using the largest possible source domain we also recommend a leave-one-out split which uses some of the target domains in the source domain. Our recommended leave-one-out splits are shown in \textbf{\color{blue}blue} in Table~\ref{tab:incrementalcrossdomain}.

\subsection{State-of-the art comparison}

In Table~\ref{tab:incrementalcrossdomain}, we also compare our multi-source, cross-domain baseline to existing cross-domain approaches, domain adaptation approaches, unsupervised approaches and supervised approaches. While direct comparison is not possible across all approaches due to the varying amount of information used during training, we can get a sense of how effective the different approaches are. 

As can be seen from Table~\ref{tab:incrementalcrossdomain}, the cross-domain methods have large variations in terms of the source domains and target domains used. Again, this is one of the motivations for this study -- establishing a standard multi-source domain and multi-target domain split. Our proposed fixed split multi-source domain (S+MT+A+3D in Table~\ref{tab:incrementalcrossdomain}) performs well compared to the existing cross-domain methods, making it a strong baseline for evaluating cross-domain approaches in future. Further improvements can be obtained using our proposed leave-one-out split.

Relative to domain adaptation methods which use unlabeled data from the target domain, our proposed multi-source domain baseline -- without any target domain information -- performs well. This also means that using multi-source domain can boost domain adaptation methods as they also use an annotated source domain.

Unsupervised methods only use unlabeled target domain information. Thus, while it is true that they benefit from information obtained from target domain, they have no manual annotation at all making these approaches difficult. Regardless, they do perform well relative to our multi-source domain baseline. However, as stated before, these methods assume the presence of training hardware at all deployment sites making these approaches less practical for real-world deployment.

Fully supervised approaches benefit from the most exact information in the form of manually annotated target domain data and can be used as a upper bound for other approaches. Relative to this upper bound, there are still considerable improvements possible for all other approaches. With our proposed fixed split and leave-one-out split multi-source domains and baseline performance, the ability to compare cross-domain approaches becomes easier and we hope to promote further investigations and improvements in cross-domain approaches to achieve similar results as the supervised approaches.

\section{Conclusion}
In this study, we established a large baseline for source and target domains to act as a comprehensive benchmark for the purpose of cross-domain person ReID.  We established the proposed baseline by conducting a systematic and comprehensive analysis of the similarities between various domains commonly used for cross-domain person ReID.  Furthermore, we investigated in an empirical fashion the effects of incrementally increasing the size of the source domain on target domain performance.  Based on our analysis and experiments, we determined and proposed a fixed and leave-one-out multi-source domain splits that performs well for six different target domains.  In addition, we established a strong baseline method for cross-domain person ReID as a benchmark reference based on lessons from past state-of-the-art methods.  We hope this study will act as a strong baseline for benchmarking to facilitate both improving and evaluating future developments in cross-domain person ReID.


{\small
\bibliographystyle{ieee}
\bibliography{egbib.bib}
}

\end{document}